\documentclass[letterpaper, 10 pt, conference]{ieeeconf}  

\IEEEoverridecommandlockouts                              

\usepackage[dvipsnames]{xcolor}

\newcommand{\ignore}[1]{}

\newcommand{\ba}{\begin{array}}
\newcommand{\ea}{\end{array}}
\newcommand{\bc}{\begin{center}}
\newcommand{\ec}{\end{center}}
\newcommand{\be}{\begin{enumerate}}
\newcommand{\ee}{\end{enumerate}}
\newcommand{\bea}{\begin{eqnarray}}
\newcommand{\eea}{\end{eqnarray}}
\newcommand{\beas}{\begin{eqnarray*}}
\newcommand{\eeas}{\end{eqnarray*}}
\newcommand{\beq}{\begin{equation}}
\newcommand{\eeq}{\end{equation}}
\newcommand{\bfig}{\begin{figure}}
\newcommand{\efig}{\end{figure}}
\newcommand{\bi}{\begin{itemize}}
\newcommand{\ei}{\end{itemize}}
\newcommand{\bpic}{\begin{picture}}
\newcommand{\epic}{\end{picture}}
\newcommand{\btabular}{\begin{tabular}}
\newcommand{\etabular}{\end{tabular}}
\newcommand{\btable}{\begin{table}}
\newcommand{\etable}{\end{table}}

\newcommand{\es}{\vfill
                 \rule[-6mm]{170mm}{0.7mm} \\
                 \redw{{\tiny
		  \hfill S-\theslide}}
                 \end{slide}}

\newcommand{\matxx}[1]{{\mathtt #1}}
\newcommand{\vecXX}[1]{{\mathbf {#1}}}
\newcommand{\vecYY}[1]{{\boldsymbol {#1}}}

\newcommand{\argmax}{\operatornamewithlimits{arg\ max}}

\def \hbar {{\bar{h}}}

\def \vecc {{\vecXX{c}}}

\def \vecn {{\vecXX{n}}}

\def \vecp {{\vecXX{p}}}

\def \vecr {{\vecXX{r}}}

\def \vect {{\vecXX{t}}}

\def \vecx {{\vecXX{x}}}

\def \vecz {{\vecXX{z}}}

\def \veceta   {{\vecYY{\eta}}}
\def \vecmu    {{\vecYY{\mu}}}

\def \matJ {{\matxx{J}}}

\def \matSigma  {{\matxx{\Sigma}}}
\def \matLambda {{\matxx{\Lambda}}}



\usepackage{amsmath,mleftright, amssymb}
\usepackage{mathtools}
\usepackage{xparse}
\usepackage{algorithm}
\usepackage[noend]{algpseudocode}
\usepackage{xcolor}
\usepackage{soul}

\newcommand{\slfrac}[2]{\left.#1\middle/#2\right.}
\usepackage{subfig}
\usepackage{graphicx}
\usepackage{float}

\NewDocumentCommand{\evalat}{sO{\big}mm}{%
  \IfBooleanTF{#1}
  {\mleft. #3 \mright|_{#4}}
  {#3#2|_{#4}}%
}

\title{\LARGE \bf
Incremental Abstraction in Distributed Probabilistic SLAM Graphs
}

\author{Joseph Ortiz, Talfan Evans, Edgar Sucar and Andrew J. Davison%
\thanks{Robot Vision Laboratory, Imperial College London, United Kingdom.}%
\thanks{{\tt\small j.ortiz@imperial.ac.uk}}%
}

\begin{document}

\maketitle
\thispagestyle{empty}
\pagestyle{empty}

\begin{abstract}

Scene graphs represent the key components of a scene in a compact and semantically rich way, but are difficult to build during incremental SLAM operation because of the challenges of robustly identifying abstract scene elements and optimising continually changing, complex graphs. We present a distributed, graph-based SLAM framework for incrementally building scene graphs based on two novel components.

First, we propose an incremental abstraction framework in which a neural network proposes abstract scene elements that are incorporated into the factor graph of a feature-based monocular SLAM system. Scene elements are confirmed or rejected through optimisation and incrementally replace the points yielding a more dense, semantic and compact representation. Second, enabled by our novel routing procedure, we use Gaussian Belief Propagation (GBP) for distributed inference on a graph processor. The time per iteration of GBP is structure-agnostic and we demonstrate the speed advantages over direct methods for inference of heterogeneous factor graphs. We run our system on real indoor datasets using planar abstractions and recover the major planes with significant compression.

\end{abstract}

\section{Introduction}

Abstract scene graphs of environments represent the key structures, objects and interactions in a semantically rich and compact way.
Ideally, an intelligent embodied device should build a scene graph rapidly and on-the-fly in a new environment using on-board sensing and processing to enable immediate intelligent action.
Identifying high-level abstractions is challenging and can require an expensive search-and-test over both the type of abstraction and the subset of elements to which it applies. Pre-trained neural networks can amortize this cost by directly proposing candidate abstractions and most algorithms for scene graph construction operate by either post-processing a low-level representation 
\cite{McCormac:etal:ICRA2017}
or by committing to abstractions of the low-level data at measurement time
\cite{Salas-Moreno:etal:CVPR2013}\cite{Sucar:etal:3DV2020}
.


\bfig
    \centering
    \includegraphics[width=\linewidth]{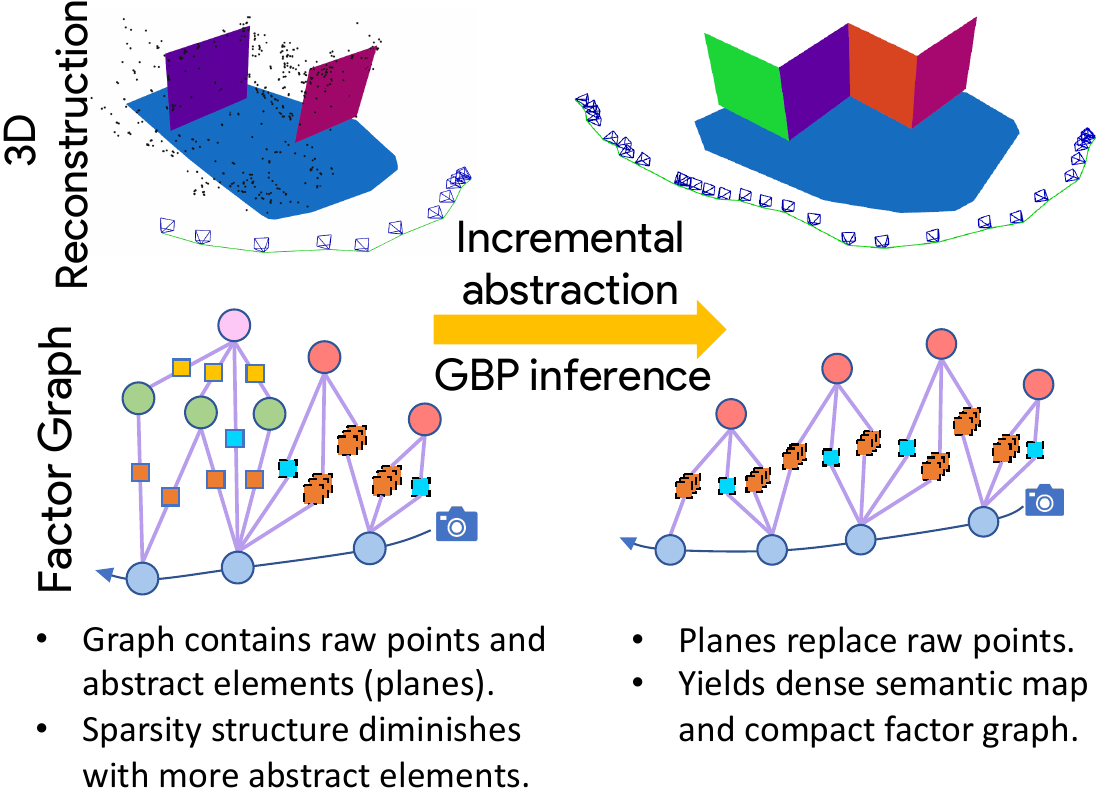}
    \caption{\textbf{Method Overview}. The time per iteration of Gaussian Belief Propagation (GBP) is structure-agnostic so it can rapidly optimise factor graphs with little sparsity structure that appear in incremental scene abstraction. In the graph, green nodes are points and red nodes are planes. See the key in Fig {\ref{fig:plane_factor_graph}} for the remaining nodes.}
    \vspace{-4mm}
    \label{fig:teaser}
\efig

A more ambitious target is general {\em incremental abstraction}.
Where abstract scene elements can be identified from single observations, they should be immediately added to a scene representation.
More commonly, several observations may be needed to identify abstractions with high confidence, requiring the system to temporarily store low-level information (e.g. raw geometry as point clouds). 
As exploration continues, abstraction should operate continually and hierarchically on both stored and incoming observations, gradually replacing the raw elements in the map. A scene graph should therefore at any point in time be a hybrid mix of raw and abstract elements, potentially of many kinds.

The correct way to accumulate many different measurements and priors into coherent estimates is via probabilistic inference, and a factor graph represents the probabilistic structure of inference problems in SLAM \cite{Dellaert:Kaess:Foundations2017}. Abstract scene elements can be combined into this estimation framework and probabilistic inference can refine and confirm or reject the abstractions. 
A hybrid, incrementally abstracting map is therefore represented by a complicated, heterogeneous and dynamically changing factor graph, where new raw structure is 
continually added while abstractions are tested, and replace raw structure if those tests are passed.

There have been few attempts to solve the true, complex inference problems these factor graphs represent in real-time systems. 
Most SLAM systems that go beyond sparse point cloud processing make severe approximations to the true inference problem by artificially layering estimation 
\cite{McCormac:etal:ICRA2017}, baking in specific variable orderings \cite{Zhou:etal:ISMAR2020}, or by using alternation to avoid joint estimation \cite{Newcombe:etal:ISMAR2011}. 
We believe these choices are often related to the rigidity of existing optimisation algorithms that need to exploit the problem structure to achieve efficient performance on standard processing hardware.

Gaussian Belief Propagation (GBP) has recently been proposed as a strong candidate algorithm for real-time inference of arbitrary and dynamically changing factor graphs \cite{Davison:Ortiz:ARXIV2019}\cite{Ortiz:etal:ARXIV2021}.
Its computational structure is node-wise parallel and it operates by local message passing on a factor graph. GBP trades the optimality of global updates for more flexible distribution of compute and memory, meaning it can better exploit parallel hardware and operate without assumptions about the global structure of an estimation problem. When implemented on a graph processor \cite{Graphcore}, GBP has already been shown to have speed advantages over global methods for inference on static bundle adjustment graphs \cite{Ortiz:etal:CVPR2020}. 

In this work, we present a general method for incrementally constructing scene graphs in real-time based on two novel components:
1) our \textbf{incremental abstraction framework} and 2) \textbf{distributed optimisation via GBP}.

First, our incremental abstraction framework combines amortized inference from an off-the-shelf network with probabilistic inference to robustly identify abstract scene elements. Additionally, upon accepting a scene element we linearise and join factors connecting common nodes to yield a more semantic, dense and compressed representation. 

Second, enabled by our novel routing procedure, we are the first to use GBP on a graph processor for inference of dynamic factor graphs. The time per iteration of GBP is independent of the graph structure and we demonstrate the resulting advantages over direct methods.

While our framework is general and can be used with any abstract feature detector, we experiment with planar abstractions, which provide a compact way to densely represent geometry in many human-made environments. 
In evaluations, our framework reconstructs accurate planar scene graphs of real indoor environments at different levels of granularity, demonstrates significant graph compression and improves tracking compared to ablated baselines.


\section{Related work}

For long-term SLAM operation it is vital to manage computational cost by limiting factor graph growth to match the spatial extent of the map. Towards this goal,  \cite{Johannsson:etal:ICRA2013} reuses existing keyframes for new measurements and \cite{Ila:etal:TRO2009} uses filtering to add only non-redundant nodes and edges. Alternatively, compression by node removal \cite{Carlevaris:etal:TRO2014}\cite{Mazuran:etal:IJRR2016}, attempts to recover the best non-linear and sparse factor graph that approximates the marginalised distribution. These methods target compression with minimal information loss, while we focus on semantic-guided compression.


Recent SLAM research has investigated incremental scene reconstruction with semantic elements included in the factor graph. 
Object-centric SLAM methods use object recognition and correspondence in the front-end to build scene graphs of objects \cite{Salas-Moreno:etal:CVPR2013}\cite{Sucar:etal:3DV2020}. Planar SLAM methods operate similarly, either detecting planes from an RGB-D input \cite{Salas-Moreno:etal:ISMAR2014}\cite{Kaess:ICRA2015}\cite{Hsiao:etal:ICRA2018}\cite{Hosseinzadeh:etal:ACCV2018}\cite{Zhou:etal:ISMAR2020}, or in the monocular case from CNN predictions \cite{Yang:etal:IROS2016}{\cite{Yang:Scherer:RAL2019}}\cite{Hosseinzadeh:etal:ICRA2019} or from the reconstructed points \cite{Arndt:etal:IROS2020}. All previous methods rely on accurate initial plane predictions and, unlike our method, cannot confirm or reject planes within the inference process, nor compress the factor graph. Additionally, for real-time operation, these methods often layer optimisation \cite{Yang:etal:IROS2016} or construct reduced systems to leverage specific problem structure \cite{Zhou:etal:ISMAR2020}.

Recent methods have built hierarchical 3D scene graphs containing layers of objects, rooms and buildings in which edges describe non-probabilistic relations \cite{Armeni:etal:ICCV2019}\cite{Rosinol:etal:ARXIV2020}\cite{Wald:etal:CVPR2020}\cite{Wu:etal:CVPR2021}. 
\cite{Armeni:etal:ICCV2019} constructs the graph from an annotated mesh model while \cite{Rosinol:etal:ARXIV2020} is an incremental system that also tracks humans.

\section{Preliminaries}


\textbf{Factor graphs} $G=(X,F,E)$ are bipartite graphs that consist of variable nodes $X=\{\vecx_i\}_{i=1:N_v}$, factor nodes $F=\{f_s\}_{s=1:N_f}$ and edges $E$. They are commonly used as a graphical representation of the factorisation of probability distributions. Given the factorisation:
$
p(X) \propto \prod_{s=1}^{N_f} f_s(X_s)
$
where $X_s \subset X$, the factor graph is constructed by creating nodes for all variables and factors and then connecting each factor $f_s$ to variables $\vecx_i \in X_s$ with an undirected edge. 



\textbf{Belief propagation} (BP) \cite{Pearl:book1988} is a distributed inference algorithm that infers the marginal distribution $p(\vecx_i)$ for each variable from the joint distribution $p(X)$. On each iteration, all factor nodes $f_s$ send messages $\mu^{t}_{f_s\rightarrow \vecx_i}$ to neighbouring variable nodes $\vecx_i \in X_s$. These messages are aggregated to update all variable node beliefs $b_i(\vecx_i)$ before all variable nodes send messages $\mu^{t}_{\vecx_i \rightarrow f_s}$ to neighbouring factor nodes. BP only involves node-wise local compute and message passing, making it trivial to distribute on highly parallel hardware by placing nodes on separate cores. Factor to variable messages are computed as:
\beq
    \label{eqn:factor_to_var_message}
    \mu^{t}_{f_s\rightarrow \vecx_i}(\vecx_i) = 
    \sum_{X_s \setminus \vecx_i} f_s(X_s) \prod_{j, \vecx_j \in X_s \setminus \vecx_i} 
    \mu^{t-1}_{\vecx_j \rightarrow f_s}(\vecx_j)
    ~,
\eeq
\noindent where $X_s \setminus \vecx_i$ denotes all elements in the set $X_s$ apart from $\vecx_i$. Variable to factor messages are computed as:
\beq
    \label{eqn:var_to_factor_message}
    \mu^{t}_{\vecx_i \rightarrow f_s}(\vecx_i) = 
     \frac{b_i^{t}(\vecx_i)}{\mu^{t-1}_{f_s\rightarrow \vecx_i}(\vecx_i)}
    ~,
\eeq
and variable beliefs are updated by simply taking the product of incoming messages from adjacent factors: $b_i^t(\vecx_i) = \prod_{l\in n(\vecx_i)} \mu^t_{f_l\rightarrow \vecx_i}(\vecx_i)$.
See \cite{Bishop:Book2006} for a full derivation. 

\subsection{GBP for Spatial Inference Problems}

BP is exact for trees, but in general it is not guaranteed to converge on the true marginals when applied to graphs with loops. Nonetheless, empirical results strongly suggest that when all distributions are Gaussian, BP performs robustly across a range of inference tasks
\cite{Bickson:PhDThesis:2008}\cite{Davison:Ortiz:ARXIV2019}.
In our GBP framework, we utilise the Gaussian information form:
$
\mathcal{N}^{-1}(\vecx;\veceta, \matLambda) \propto \; \exp{(-\frac{1}{2} \vecx^\top \matLambda \vecx + \veceta^\top \vecx)}
~,
$
where $\veceta = \Sigma^{-1} \vecmu$ is the information vector and $\matLambda = \Sigma^{-1}$ the precision matrix.

Measurements are modelled as a deterministic function plus centred Gaussian noise, $Z = h(X) + \epsilon$, $\epsilon \sim \mathcal{N}(0, \matSigma_M)$. 
In spatial estimation problems independent measurements $z_m$ often depend on only a small subset of the variables $X_m$. The likelihood therefore has the factorised form:
\beq
    \small
    \label{eqn:factorised_likelihood}
    l(X;Z) = 
    \prod_{m}^{N_m} l_m(X_m; z_m) \propto 
    \prod_{m}^{N_m}
    \exp(-\frac{1}{2} 
    \parallel h_m(X_m) - z_m \parallel^2_{\matSigma_m})
    ~,
\eeq    
and the factor graph for the posterior $P(X|Z)$ consists of a variable node for each $\vecx_i$ and a factor node for each $l_m$.

We are interested in estimating both the configuration $X$ that maximises the posterior distribution $P(X|Z)$ given observed measurements $Z$ and the associated uncertainty given by the marginal covariances. Using Bayes Rule:
\beq
    \small
    \label{eqn:map}
    X^{MAP} = \argmax_{X} p(X | Z) = \argmax_{X} l(X ; Z) p(X)    
    ~,
\eeq
where $l(X; Z)$ is proportional to $P(Z|X)$ but is a function of the variables $X$ with the measurements as parameters \cite{Dellaert:Kaess:Foundations2017}.


Gaussian Belief Propagation performs posterior inference on the factor graph by computing the marginal posteriors $p(\vecx_i \rvert Z) =  \mathcal{N}(\mu_i, \Sigma_i)$ for all variables, where $X^{MAP} = [\mu_1, ... , \mu_{N_v}]^{\top}$. 
For factors with non-linear measurement functions $h$, the density $l$ is non-Gaussian and so we use the first order Taylor expansion, $h(X) \approx h(X_{0}) + J(X - X_{0})$, to yield a Gaussian likelihood (dropping $m$ subscripts) \cite{Davison:Ortiz:ARXIV2019}:
\beq
    \small
    \label{eqn:gaussian_linearisation}
    l(X; z) = \mathcal{N}^{-1}(X; 
    \matJ^\top \matSigma^{-1} ( \matJ\; X_0 + z - h(X_0) ), 
    \matJ^\top \matSigma^{-1} \matJ
    )
    ~.
\eeq
During inference, non-linear factors are relinearised independently when the $L1$ distance between the current beliefs and the linearisation point exceeds a threshold $\beta$.


\section{Incremental Planar Abstraction Framework}

We combine amortized inference via a neural network with distributed probabilistic inference via GBP to incrementally abstract factor graphs in SLAM. The master scene representation is always the factor graph and during online operation GBP is continually performing inference on this graph. 

GBP is interrupted to edit the factor graph, which occurs when: i) adding a new keyframe (Sec. \ref{sec:raw_front_end}, \ref{sec:plane_pred_filtering}) ii) testing plane hypotheses (Sec. \ref{sec:confirm_reject}) or iii) merging planes (Sec. {\ref{sec:merge}). An overview of the system is provided in Algorithm \ref{alg:system}.

\begin{algorithm}[b]
\caption{System Overview.}
\label{alg:system}
\begin{algorithmic}[1]
    \small
    \State Initialise factor graph with first keyframe, n\_iterations = 0
    \While {in operation}
        \State Run iteration of GBP,  n\_iterations += 1
        \If {new keyframe}
            \State Feature matching (Sec. \ref{sec:raw_front_end},  Fig. \ref{fig:plane_factor_graph} \textit{left})
            \State Add NN plane hypotheses (Sec. \ref{sec:plane_pred_filtering}, Fig. \ref{fig:plane_factor_graph} \textit{mid})
        \EndIf
        \If {n\_iterations \% N == 0}
            \For {each plane hypothesis}
                \If {confirm criteria satisfied (Eq. \ref{eqn:confirm_criteria})}
                    \State Create rigid body plane (Fig. \ref{fig:plane_factor_graph} \textit{right})
                \ElsIf {reject criteria satisfied (Eq. \ref{eqn:reject_criteria})}
                    \State Remove plane hypothesis (Fig. \ref{fig:plane_factor_graph} \textit{left})
                \EndIf
            \EndFor
        \EndIf
        \If {n\_iterations \% M == 0}
            \For {each pair of planes}
                \If {merge criteria satisfied }
                    \State Merge planes (Sec. \ref{sec:merge})
                \EndIf
            \EndFor
        \EndIf
    \EndWhile
\end{algorithmic}
\end{algorithm}

\subsection{Feature Extraction from Keyframes \label{sec:raw_front_end}}

We construct the abstract scene graph on top of a feature-based monocular SLAM system. 
We choose a sparse front-end for experimental purposes and concentrate on the graphical back-end in this paper.
Given a stream of live images, we use the ORB-SLAM2 \cite{Mur-Artal:etal:TRO2017} front-end to build the factor graph composed of keyframes, points and reprojection factors (Fig. \ref{fig:plane_factor_graph} \textit{left}). 
Reprojection factors penalise the distance between a matched feature at image coordinates $\vecz$ in keyframe $\vecc$ and the projection of the corresponding point $\vecp$ in the image plane:
\beq
    \label{eqn:reprojection_hfunc}
    \small
    l_r (\vecc, \vecp\,; \vecz) \propto \exp \big( -\frac{1}{2} \parallel \vecz - \text{proj} \, (R_c \, \vecp + \vect_c) \parallel^2_{\matSigma_{r}} \big) 
    ~,
\eeq
where $\text{proj}$ is the projection operator and $R_c$ and $\vect_c$ are the rotations and translations derived from keyframe pose $\vecc$.

\begin{figure*}
    \vspace{0.5em}
    \centering
    \includegraphics[width=\linewidth]{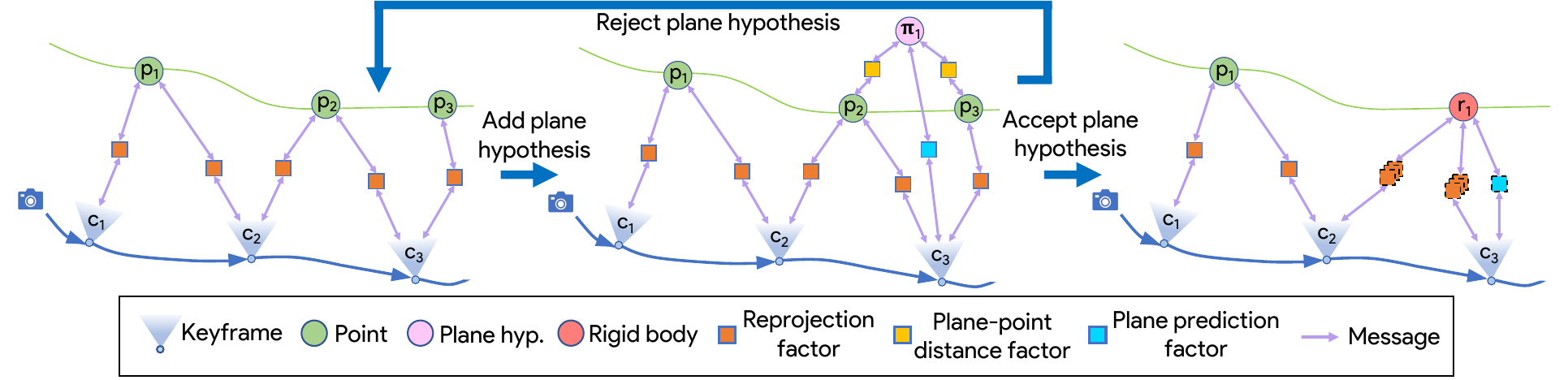}
    \caption{\small \textbf{Factor Graph.} \textit{Left}: Raw factor graph with only keyframes and points. \textit{Middle}: The network predicts plane $\boldsymbol{\pi}_1$ in keyframe 3 and points $\vecp_2$ and $\vecp_3$ lie inside the predicted segmentation mask. The plane hypothesis variable node is added to the graph along with 2 plane-point distance factors and a plane prediction factor. \textit{Right}: The plane hypothesis is confirmed and the plane hypothesis and points are replaced by a rigid body plane node $\vecr_1$. The factors with dashed borders connect to a rigid body node and have a different functional form (Eq. \ref{eqn:plan_pred_hfunc} and \ref{eqn:reprojection_hfunc_v2}). Linearised reprojection factors that connect to the same rigid body node are combined into a single factor which is represented by overlaying 3 reprojection factors. If the plane hypothesis is rejected the factor graph returns to the form on the left.}
    \vspace{-4mm}
    \label{fig:plane_factor_graph}
\end{figure*}

\subsection{Integrating Plane Predictions \label{sec:plane_pred_filtering}}

We denote the homogeneous plane vector $\boldsymbol{\pi} = (\pi_1, \pi_2, \pi_3, \pi_4)^{\top} \in \mathbb{P}^3$ \cite{Hartley:Zisserman:Book2004}. Points $\vecp \in \mathbb{R}^3$ lying in a plane satisfy $\hat{\vecn}^{\top} \vecp = d$, where $\hat{\vecn} = \frac{(\pi_1, \pi_2, \pi_3)^{\top}}{\sqrt{\pi_1^2 + \pi_2^2 + \pi_3^2}}$ is the normal vector and $d = \frac{-\pi_4}{\sqrt{\pi_1^2 + \pi_2^2 + \pi_3^2}}$ is the distance from the origin. The homogeneous plane vector is transformed between coordinate frames using the inverse transpose of the homogeneous point transform: $\boldsymbol{\pi}^{\prime} = T^{-\top} \boldsymbol{\pi}$. For optimisation, we represent planes using the minimal parametrisation $\hat{\vecn} \cdot d$ and denote the $\boxminus$ operator to subtract the plane parameters in our chosen minimal form ($\boldsymbol{\pi_a} \boxminus \boldsymbol{\pi_b} \coloneqq \hat{\vecn}_a \cdot d_a - \hat{\vecn}_b \cdot d_b)$.

For each keyframe, we run a forward pass of the PlaneRCNN model \cite{Liu:etal:CVPR2019} to predict a set of plane parameters and corresponding segmentation masks. The model is based on Mask RCNN {\cite{He:etal:ICCV2017} with an additional head to regress the plane normal. The predictions are filtered to remove planes with small or disconnected segmentation masks. The resulting {\em plane hypotheses} are then integrated into the existing factor graph as plane hypothesis nodes (Fig. \ref{fig:plane_factor_graph} \textit{middle}). 

Using the segmentation mask for each plane, we determine the map points that are predicted to lie in the plane and introduce plane-point distance factors, connecting the hypothesised plane $\boldsymbol{\pi}$ to each of these points $\vecp$. Plane-point distance factors penalise the perpendicular distance from a point to a plane and have the form: 
\beq
    \label{eqn:plane_to_point_hfunc}
    \small
    l_{pp} (\vecp, \boldsymbol{\pi}) \propto \exp \big( -\frac{1}{2} \parallel \hat{\vecn} \cdot \vecp - d  \parallel^2_{ \matSigma_{pp} } \big) 
    ~.
\eeq
Each plane hypothesis $\boldsymbol{\pi}$ is also connected to the keyframe $\vecc$ in which it was predicted via a plane prediction factor. Plane prediction factors treat the network prediction of the plane parameters $\boldsymbol{\pi}_z$ as a measurement and 
take the form:
\beq
    \label{eqn:plane_hyp_pred_hfunc}
    \small
    l_{\pi p } (\boldsymbol{\pi}, \vecc \, ; \boldsymbol{\pi}_z) \propto \exp
    \big( -\frac{1}{2} \parallel \boldsymbol{\pi}_z 
    \boxminus
    T_{cw}^{-\top} \boldsymbol{\pi}  \parallel^2_{ \matSigma_{\pi p} } \big) 
    ~,
\eeq
where $T_{cw} \in SE(3)$ is the transformation from the global coordinate frame to the coordinate frame of the camera $\vecc$. In experiments we set $\matSigma_{\pi p}$ to be very large as the network plane parameter prediction can be unreliable.

Our framework is not specific to planes; in fact, any abstract scene element with appropriate compatibility factor and inference model to generate hypotheses could be used.

\subsection{Plane hypothesis confirmation and rejection \label{sec:confirm_reject}}

Having added the plane hypotheses to the raw factor graph, GBP carries out inference on the hybrid graph (Fig. \ref{fig:plane_factor_graph} \textit{middle}) and converges to the configuration that minimises the factor energies. To allow bad plane hypotheses to be treated as outlying measurements and only contribute weakly to the graph energy, we employ the robust Tukey loss function for all factors via covariance rescaling as in \cite{Davison:Ortiz:ARXIV2019} and \cite{Agarwal:etal:ICRA2013}.

After convergence, for each plane hypothesis, we go through all connected points and read off from the factor graph the likelihood that the point lies in the plane. The likelihood is the plane-point distance factor density $l_{pp}(\vecp_{conv}, \boldsymbol{\pi}_{conv})$ evaluated at the converged belief means $\vecp_{conv}$ and $\boldsymbol{\pi}_{conv}$. To determine whether to confirm or reject each plane hypothesis, we use the proportion of points $y$ with likelihood $l_{pp}(\vecp_{conv}, \boldsymbol{\pi}_{conv}) > l_{thresh}$ and the number of iterations the hypothesis has been in the graph $t$.
\begin{align}
    \small
    & \textbf{Reject criteria}: \;\;y < y_{reject} \:\:\; \textit{OR} \:\:\; t >t_{max} \label{eqn:reject_criteria}\\
    & \textbf{Confirm criteria}: \;\; y >  y_{conf} \:\:\; \textit{AND} \:\:\;  t >t_{min} \label{eqn:confirm_criteria}
\end{align}

If rejected, the plane hypothesis node and all adjacent factors are removed from the graph. If confirmed, the plane hypothesis node and points with $l_{pp}(\vecp_{conv}, \boldsymbol{\pi}_{conv}) > l_{thresh}$ are replaced by a single rigid body variable node with just 6 degrees of freedom, as in Fig. \ref{fig:plane_factor_graph} \textit{right}. We use a rigid body to represent the plane as we assume that the relative configuration of the planar points has now been determined and that they can be optimised as a single rigid planar body.

The reprojection factors connected to the planar points, and the plane prediction factors connected to the plane hypothesis node are transferred to the new rigid body variable node. 
For these factors, the rigid body transformation $\vecr$ becomes an argument of the measurement function while $\boldsymbol{\pi}_{conv}$ and $\vecp_{conv}$ are parameters as we replace:
\beq
    \label{eqn:accepting_plane}
    \small
    \boldsymbol{\pi} \rightarrow T_r^{-\top} \boldsymbol{\pi}_{conv} \;\;\; \text{and} \;\;\;
    \vecp \rightarrow R_r \vecp_{conv} + \vect_r
    ~,
\eeq
where $T_r$, $R_r$ and $\vect_r$ are the transformation matrix, rotation matrix and translation derived from the minimal $SE(3)$ vector $\vecr$. Plane prediction and reprojection factors become:
\beq
    \small
    \label{eqn:plan_pred_hfunc}
    l_{\pi p} (\vecr, \vecc ; \boldsymbol{\pi}_z, \boldsymbol{\pi_{conv}}) \propto \exp \big( -\frac{1}{2} \parallel \boldsymbol{\pi}_z \boxminus T_{cw}^{-\top} T_{r}^{-\top} \boldsymbol{\pi}_{conv}  \parallel^2_{ \matSigma_{\pi p} } \big)
    ~,
\eeq
\beq
    \small
    \label{eqn:reprojection_hfunc_v2}
    l_r (\vecc, \vecr; \vecz, \vecp_{conv}) \propto \exp \big( - \slfrac{1}{2} \parallel \vecz - proj  (R_c  (R_r \vecp + \vect_r) + \vect_c) \parallel^2_{\matSigma_{r}} \big)
\eeq

As all transferred reprojection factors connect to the same pair of nodes, they are linearised and combined into a single factor by taking the product.
This further compresses the graph and reduces the computation at each GBP iteration. 
Note that relinearisation requires linearising the contributions from all the original reprojection factors, however this only occurs on a small proportion of GBP iterations.

\subsection{Merging Planes \label{sec:merge}}

As there may be multiple plane hypotheses for the same geometric plane, we merge confirmed planes that have: i) aligned normal vectors ii) small perpendicular separation and iii) large overlap. The latter two criteria are estimated by sampling points in the plane. Once two planes are chosen to be merged, we replace the two rigid body planes with a single rigid body plane. The new plane normal is the average of the merged normals and the factors connecting to the merged planes are transferred to the new plane. 

\section{Dynamic Routing on the IPU}

We implement our incremental abstraction method on a single Graphcore MK1 IPU chip \cite{Graphcore} as in \cite{Ortiz:etal:CVPR2020}. The IPU is a large chip composed of 1216 independent cores arranged in a fully connected graph structure. Like a GPU it is highly parallel, but due to its interconnect structure and the local memory on each core, it has breakthrough performance for algorithms with a sparse message passing character. 

Inference on static factor graphs is achieved in \cite{Ortiz:etal:CVPR2020} by mapping the factor graph onto the IPU cores with the communication pattern matching the topology of the graph. This approach however cannot be applied to dynamic graphs, as the communication pattern must be precompiled and recompilation is expensive. To enable parallel inference of dynamic factor graphs, we develop a routing procedure that can manage arbitrary graph topologies while operating on a fixed precompiled communication pattern.

Our routing solution introduces densely connected routing nodes to mediate the transfer of messages through the factor graph. Routing nodes have knowledge about the structure of a part of the factor graph, stored in a routing matrix. As shown in Fig. \ref{fig:routing}, when the factor graph is edited, the routing matrix can be updated to enable inference on the new factor graph without changing the compiled communication pattern.

To distribute the routing, we create a routing node for each factor type. One requirement is to specify the maximum number of nodes of each type in the graph and the maximum number of edges each type of node can have -- these are weak requirements and the limits can be set generously. For large graphs we create multiple routing nodes for each factor type.

\bfig
    \vspace{0.5em}
    \centering
    \includegraphics[width=\linewidth]{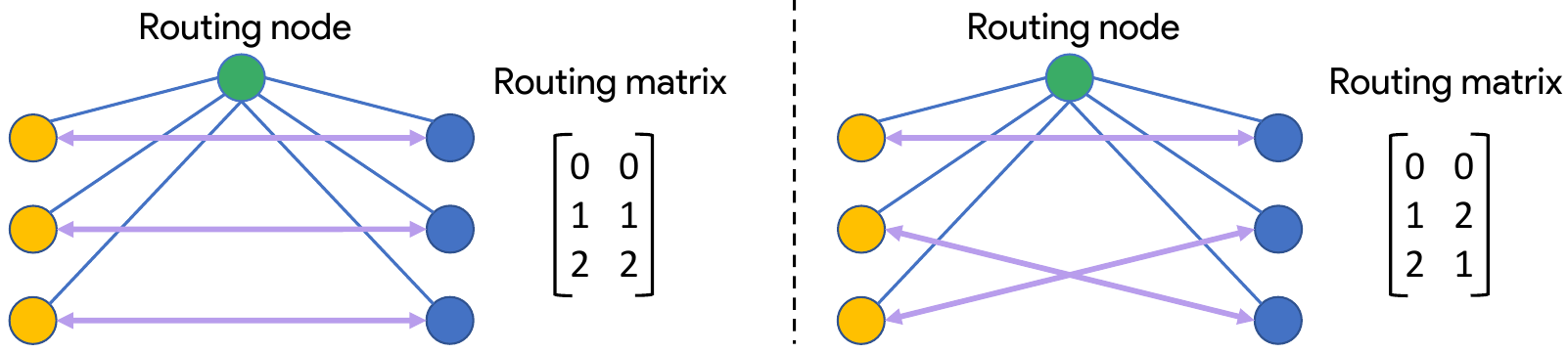}
    \caption{\textbf{Routing procedure}. Yellow and blue nodes are different types of node. Purple lines represent edges in the factor graph and blue lines communication edges between hardware cores. In this illustration, each node in the graph lives on a different core. }
    \vspace{-4mm}
    \label{fig:routing}
\efig

The routing procedure enables optimisation of arbitrary graph topologies at minimal time cost; only increasing the time per iteration from $200\mu s$ to $400 \mu s$. 
As GBP typically converges in less than 100 iterations \cite{Ortiz:etal:CVPR2020}, inference remains comfortably within real-time constraints.

\section{Experimental Evaluation \label{sec:evaluation}}

For evaluations, we use real-world sequences captured with the Kinect camera in varied indoor environments and sequences from the TUM dataset \cite{Sturm:etal:IROS2012}. We compare against \cite{Ortiz:etal:CVPR2020} which uses GBP for inference but without planar abstractions. We call this method GBP-BL as it is similar to our system but without planes. 


\textbf{Implementation settings.}
For all experiments, we use the default parameters: $\Sigma_r = \sigma_r^2 I_2$, $\sigma_r = 2$ pixels, $\Sigma_{\pi p} = \sigma_{\pi p}^2 I_3$, $\sigma_{\pi p} = 20m$, $\Sigma_{pp} = \sigma_{pp}^2$, $\sigma_{pp} = 5cm$, $y_{reject} = 0.5$, $y_{conf}=0.8$, $l_{thresh}=0.8$, $t_{max}=6000$ iterations,  $t_{min}=4000$ iterations, $N = M = 1000$ iterations and $\beta = 1e-4$. 
We use message dropout of $0.7$ and message damping of $0.4$ to stabilise GBP \cite{Bickson:PhDThesis:2008}. 
Keyframes are initialised with a constant velocity motion model and points using an average depth.

\subsection{Convergence Time Evaluation \label{sec:speed}}

\begin{figure}[t]
    \centering
    \includegraphics[width=\linewidth]{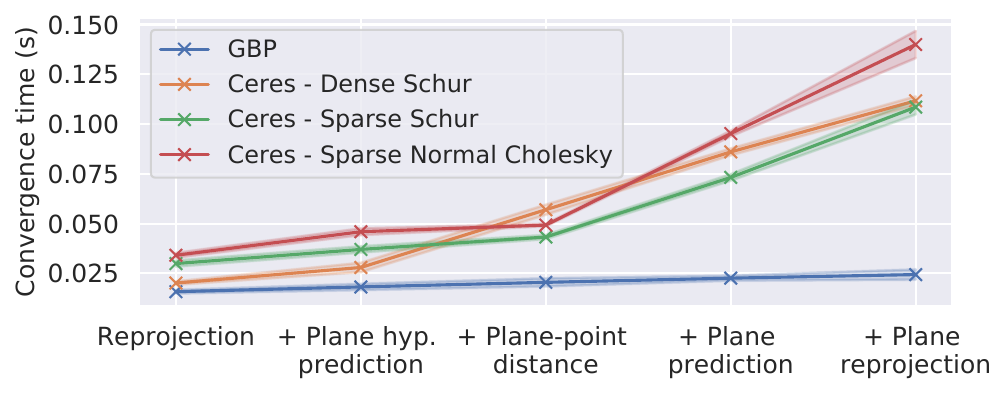}
    \caption{
    We plot mean and standard deviation error over 10 runs. Convergence is defined as reaching 1.5 pixels average reprojection error as in \cite{Ortiz:etal:CVPR2020}. 
    With the additional 800 factors, convergence time for Ceres with Dense Schur increases by roughly 5x from $20.1ms$ to $111.6ms$, while GBP increases from $15.6ms$ to $22.0ms$.}
    \vspace{-4mm}
    \label{fig:speed}
\end{figure}

\begin{figure*}[ht]
    \vspace{0.5em}
    \centering
    \includegraphics[scale=0.5]{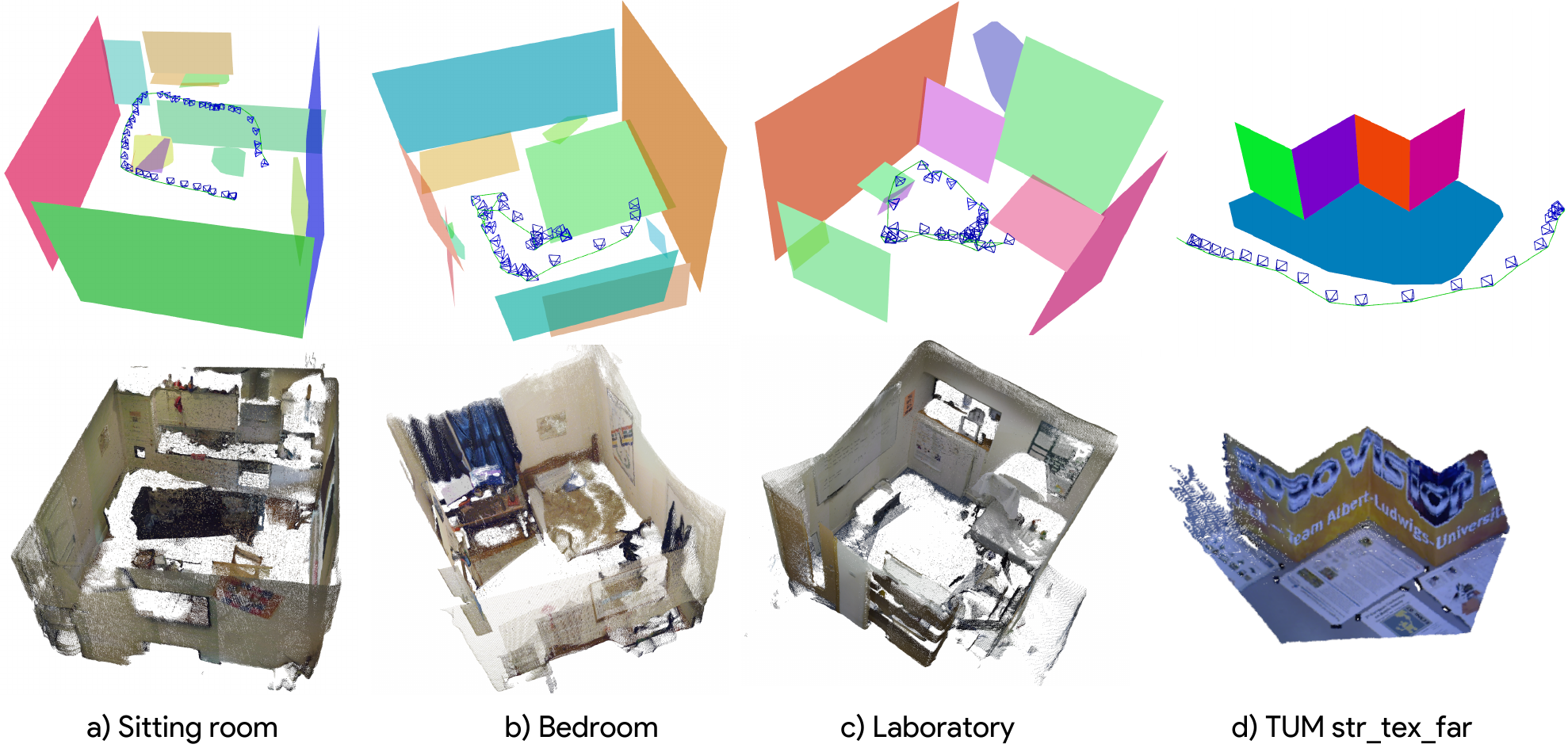}
    \caption{\textbf{Qualitative reconstructions}. \textit{Top}: Planar reconstructions recovered by our system. Planes that are almost perpendicular or parallel to the largest planes in the reconstruction are displayed as rectangles, while the extent of other planes is the convex hull of the planar points. We do not include raw points that have not been abstracted in this visualisation. \textit{Bottom}: Visualisation of the scene obtained using depth, shown for reference rather than comparison. Reconstructions \textit{a)}, \textit{b)} and \textit{c)} are from sequences captured with the Kinect camera and \textit{d)} is from the TUM sequence \textit{structure texture far}. Perpendicular planes in \textit{d)} are joined at their intersection for a watertight reconstruction. Note in sequence \textit{c)}, the near side of the room is not abstracted into a plane as there is a cluttered bookshelf.}
    \vspace{-4mm}
    \label{fig:qualitative_res}
\end{figure*}

We evaluate the convergence time for bundle adjustment problems based on the \textit{sitting room} sequence for factor graphs with an increasing number of different factor types. We begin with a factor graph containing 35 keyframes, 3108 points and 10000 reprojection factors and measure the convergence time from noisy initialisations. We then add 200 additional factors of a new type to the graph along with any new variables nodes (for example we add plane hypotheses variable nodes when adding plane hypothesis prediction factors) and repeat the experiment. We conduct experiments adding 4 additional types of factors, meaning in the final experiment there are 10800 factors in the graph.

Following \cite{Ortiz:etal:CVPR2020}, we compare GBP on a single IPU chip \cite{Graphcore} with Ceres \cite{CeresManual}, a non-linear least squares optimisation library, run on a 6 core i7-8700K CPU with 18 threads. Ceres uses Levenberg-Marquardt (LM), a Tukey kernel and analytic derivatives. In Fig. \ref{fig:speed}, we compare the convergence time of GBP with the 3 fastest Ceres linear solvers.

As different types of factors are added to the graph, convergence time for Ceres increases greatly while GBP increases only marginally.
It is instructive to break down convergence time as the product of the time per iteration and the number of iterations to converge. 
Ceres makes optimal global updates through LM and, across all experiments, converges in 5-10 iterations while GBP requires 20-25 iterations. 
The time per iteration however is the more significant factor and where the two methods differ greatly in performance.

The local distributed nature of GBP makes its time per iteration \textit{structure-agnostic}; in other words, the compute per iteration depends only on the number of factors and not their structure. 
Consequently, the time per iteration for GBP is approximately constant for all experiments and the convergence time only increases slightly, due to extra factors and more iterations required to converge.

In contrast, as different types of factors are added to the graph, solving the linear system or normal equations for the Ceres LM update becomes considerably more expensive, increasing the time per iteration. 
Ceres linear solvers compute the update by exploiting sparsity structure in the Fisher information matrix which depends on both the number of non-zero entries or equivalently factors in the graph and the variable ordering. In the base case, the Dense Schur solver is designed to leverage the large zero blocks in the information matrix to efficiently solve the normal equations without inverting the full information matrix. As different types of factors are added, even in very small numbers, these zero blocks are eroded and the time per iteration for Ceres increases by over 5x with only an additional $8\%$ of factors.

These experiments expose the reliance of direct solvers on fixed sparsity structure and suggest that GBP is more efficient for optimising heterogeneous scene graphs without strong structure.
Lastly, not only does GBP have the right computational properties, but it is also doing additional work by computing both the MAP and the marginal covariances while LM only computes the MAP with significant extra computation required to get the covariances.

\subsection{Qualitative Reconstructions}

We show planar reconstructions for 4 real sequences in Fig. \ref{fig:qualitative_res}. Our system captures the prominent planes such as walls, beds (\textit{b}), desks (\textit{b}, \textit{c}) and cupboards (\textit{a}). In Fig. \ref{fig:qualitative_res} \textit{d)} we verify that for a simple planar scene, our system can achieve a complete reconstruction. Fig. \ref{fig:teaser} shows an intermediate reconstruction of the same sequence with points.

\subsection{Compression Evaluation}

Graph compression yields a more dense, semantic and parameter-efficient representation and reduces the amount of computation per iteration of GBP. The computation is proportional to the number of edges or equivalently the number of factors when all factors are pairwise. To quantify the compression capabilities of our system, we compare the number of factor nodes in the graph with GBP-BL in Fig. \ref{fig:num_var_nodes}.

For our planar system, there are initially more factors as many plane hypotheses are added, however the graph is quickly compressed as planes are confirmed. The compression is most significant in the \textit{Laboratory} and \textit{TUM str\_tex\_far} sequences in which there are large textured walls, while there is less compression in the \textit{sitting room} sequence due to large curved objects such as the sofa. 

\bfig
    \vspace{0.4em}
    \centering
    \includegraphics[width=\linewidth]{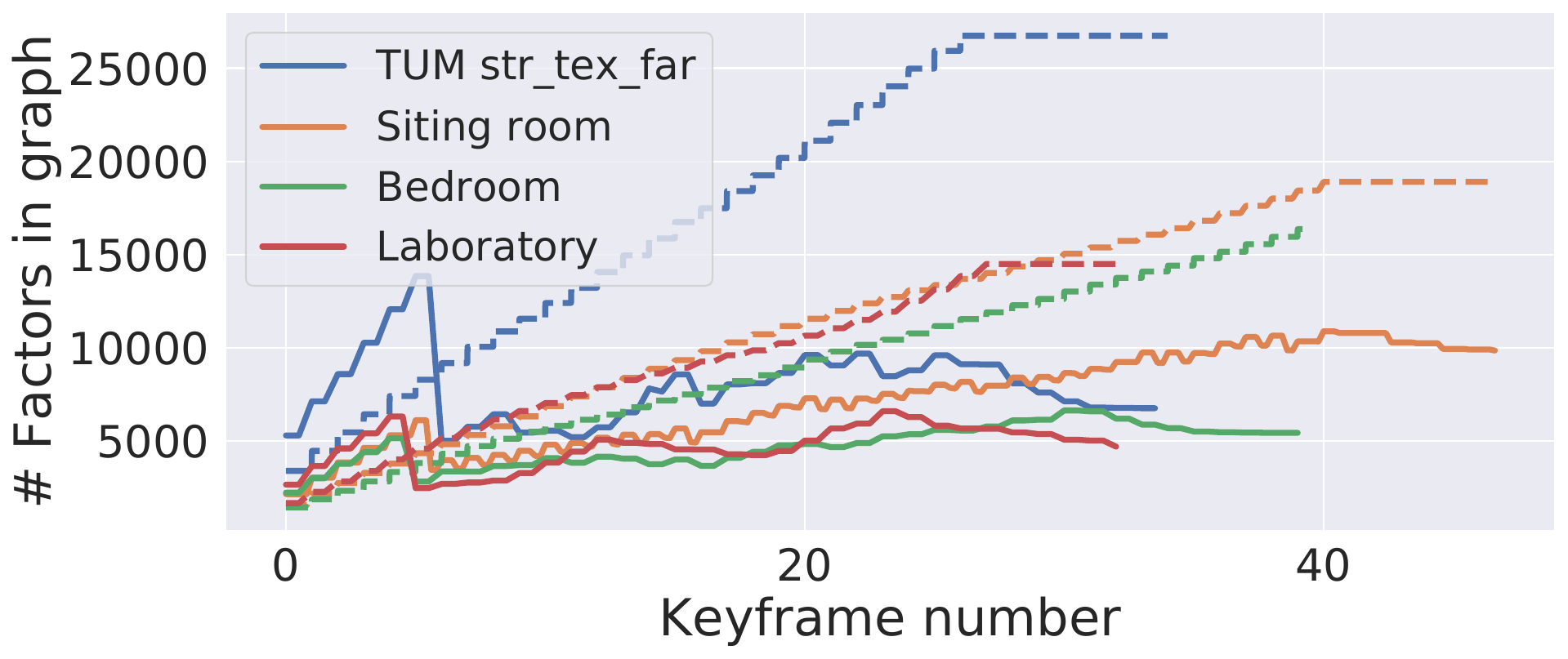}
    \caption{We compare the number of factor nodes in the graph during a sequence. Solid lines are our system, dashed lines are GBP-BL.}
    \vspace{-4mm}
    \label{fig:num_var_nodes}
\efig

\subsection{Tracking Evaluation}

\begin{table}[t]
    \vspace{0.5em}
    \centering
    \small
    \caption{Average absolute trajectory error (ATE) over 16 runs for TUM sequences \cite{Sturm:etal:IROS2012}. Ours-C is ours without compression.}
    \label{tab:ate}    
    \begin{tabular}{| c || c | c | c | c |} 
        \hline
        ATE (cm) & Ours & Ours-C & GBP-BL  & ORB-SLAM2 \\
        \hline
        str\_tex\_far & 1.204 & 1.186 & 1.384 & \textbf{0.924} \\
        \hline
        cabinet & 0.723 & 0.659 & 1.048 & \textbf{0.601} \\
        \hline
        long\_office & 0.658 & \textbf{0.648} & 0.891 & 0.670 \\
        \hline
    \end{tabular}
    \vspace{-4mm}
\end{table}

We evaluate the absolute trajectory error (ATE) on 3 TUM sequences (chosen for their prominent planar regions) in Table \ref{tab:ate}. We compare our full method with two ablated systems: our planar method without compression (Ours-C) and GBP-BL which has similar performance to Ceres \cite{Ortiz:etal:CVPR2020}. 

Ours-C has lower ATE than GBP-BL demonstrating that planar constraints help tracking.
Our full method performs slightly worse than Ours-C because compression is lossy, however the fact that the difference is small indicates that the compression is accurate. 
Our tracking is comparable to ORB-SLAM2 \cite{Mur-Artal:etal:TRO2017}, even achieving lower ATE for one sequence with slow motion and many planes. Both our method and GBP-BL lack a tracking system so receive worse initialisations than ORB-SLAM2 when new frames are added, likely explaining the worse performance on two sequences. 
Designing a distributed tracking system within the GBP framework remains an important research direction.

\section{Conclusions}

We have proposed a method for efficient incremental construction of probabilistic scene graphs from monocular input based on two novel components. 
First, our incremental scene abstraction framework combines amortized inference with probabilistic inference to identify abstract scene elements and build a semantic, dense and compact representation. 
We show that with planar abstractions, we can achieve accurate reconstructions with significant compression.
Second, our routing procedure enables inference on dynamic graphs with GBP on a graph processor. 
We demonstrate the advantage of GBP over direct methods for complex factor graphs due to the structure-agnostic time per iteration.
In the near term, we hope to inspire research into novel parallel processors to tackle the computational challenges of optimising dynamic heterogeneous graphs.

{\small
\bibliographystyle{IEEEtran}
\bibliography{IEEEabrv, robotvision}

\begin{thebibliography}{10}
\providecommand{\url}[1]{#1}
\csname url@rmstyle\endcsname
\providecommand{\newblock}{\relax}
\providecommand{\bibinfo}[2]{#2}
\providecommand\BIBentrySTDinterwordspacing{\spaceskip=0pt\relax}
\providecommand\BIBentryALTinterwordstretchfactor{4}
\providecommand\BIBentryALTinterwordspacing{\spaceskip=\fontdimen2\font plus
\BIBentryALTinterwordstretchfactor\fontdimen3\font minus
  \fontdimen4\font\relax}
\providecommand\BIBforeignlanguage[2]{{%
\expandafter\ifx\csname l@#1\endcsname\relax
\typeout{** WARNING: IEEEtran.bst: No hyphenation pattern has been}%
\typeout{** loaded for the language `#1'. Using the pattern for}%
\typeout{** the default language instead.}%
\else
\language=\csname l@#1\endcsname
\fi
#2}}

\bibitem{McCormac:etal:ICRA2017}
J.~McCormac, A.~Handa, A.~J. Davison, and S.~Leutenegger, ``{SemanticFusion}:
  Dense {3D} semantic mapping with convolutional neural networks,'' in
  \emph{{Proceedings of the {IEEE} International Conference on Robotics and
  Automation ({ICRA})}}, 2017.

\bibitem{Salas-Moreno:etal:CVPR2013}
\BIBentryALTinterwordspacing
R.~F. Salas-Moreno, R.~A. Newcombe, H.~Strasdat, P.~H.~J. Kelly, and A.~J.
  Davison, ``{{SLAM++}: Simultaneous Localisation and Mapping at the Level of
  Objects},'' in \emph{{Proceedings of the {IEEE} Conference on Computer Vision
  and Pattern Recognition ({CVPR})}}, 2013. [Online]. Available:
  \url{http://dx.doi.org/10.1109/CVPR.2013.178}
\BIBentrySTDinterwordspacing

\bibitem{Sucar:etal:3DV2020}
E.~Sucar, K.~Wada, and A.~Davison, ``{NodeSLAM}: Neural object descriptors for
  multi-view shape reconstruction,'' in \emph{Proceedings of the International
  Conference on 3D Vision ({3DV})}, 2020.

\bibitem{Dellaert:Kaess:Foundations2017}
F.~Dellaert and M.~Kaess, ``{Factor Graphs for Robot Perception},''
  \emph{Foundations and Trends in Robotics}, vol.~6, no. 1--2, pp. 1--139,
  2017.

\bibitem{Zhou:etal:ISMAR2020}
L.~Zhou, D.~Koppel, H.~Ju, F.~Steinbruecker, and M.~Kaess, ``An efficient
  planar bundle adjustment algorithm,'' in \emph{{Proceedings of the
  International Symposium on Mixed and Augmented Reality ({ISMAR})}}, 2020.

\bibitem{Newcombe:etal:ISMAR2011}
R.~A. Newcombe, S.~Izadi, O.~Hilliges, D.~Molyneaux, D.~Kim, A.~J. Davison,
  P.~Kohli, J.~Shotton, S.~Hodges, and A.~Fitzgibbon, ``{{KinectFusion}:
  Real-Time Dense Surface Mapping and Tracking},'' in \emph{{Proceedings of the
  International Symposium on Mixed and Augmented Reality ({ISMAR})}}, 2011.

\bibitem{Davison:Ortiz:ARXIV2019}
A.~J. Davison and J.~Ortiz, ``{FutureMapping 2: Gaussian Belief Propagation for
  Spatial AI},'' \emph{arXiv preprint arXiv:arXiv:1910.14139}, 2019.

\bibitem{Ortiz:etal:ARXIV2021}
J.~Ortiz, T.~Evans, and A.~J. Davison, ``A visual introduction to gaussian
  belief propagation,'' \emph{arXiv preprint arXiv:2107.02308}, 2021.

\bibitem{Graphcore}
``Graphcore,'' URL https://www.graphcore.ai/.

\bibitem{Ortiz:etal:CVPR2020}
J.~Ortiz, M.~Pupilli, S.~Leutenegger, and A.~Davison, ``{Bundle Adjustment on a
  Graph Processor},'' in \emph{{Proceedings of the {IEEE} Conference on
  Computer Vision and Pattern Recognition ({CVPR})}}, 2020.

\bibitem{Johannsson:etal:ICRA2013}
H.~Johannsson, M.~Kaess, M.~Fallon, and J.~Leonard, ``{Temporally scalable
  visual SLAM using a reduced pose graph},'' in \emph{{Proceedings of the
  {IEEE} International Conference on Robotics and Automation ({ICRA})}}, 2013.

\bibitem{Ila:etal:TRO2009}
V.~Ila, J.~Porta, and J.~Andrade-Cetto, ``{Information-based compact pose
  SLAM},'' \emph{IEEE Transactions on Robotics}, vol.~26, no.~1, pp. 78--93,
  2009.

\bibitem{Carlevaris:etal:TRO2014}
N.~Carlevaris-Bianco, M.~Kaess, and R.~Eustice, ``Generic node removal for
  factor-graph slam,'' \emph{IEEE Transactions on Robotics}, vol.~30, no.~6,
  pp. 1371--1385, 2014.

\bibitem{Mazuran:etal:IJRR2016}
M.~Mazuran, W.~Burgard, and G.~D. Tipaldi, ``Nonlinear factor recovery for
  long-term {SLAM},'' \emph{{International Journal of Robotics Research
  ({IJRR})}}, vol.~35, no.~1, pp. 50--72, 2016.

\bibitem{Salas-Moreno:etal:ISMAR2014}
R.~F. Salas-Moreno, B.~Glocker, P.~H.~J. Kelly, and A.~J. Davison, ``Dense
  planar {SLAM},'' in \emph{{Proceedings of the International Symposium on
  Mixed and Augmented Reality ({ISMAR})}}, 2014.

\bibitem{Kaess:ICRA2015}
M.~Kaess, ``Simultaneous localization and mapping with infinite planes,'' in
  \emph{{Proceedings of the {IEEE} International Conference on Robotics and
  Automation ({ICRA})}}, 2015.

\bibitem{Hsiao:etal:ICRA2018}
M.~Hsiao, E.~Westman, and M.~Kaess, ``Dense planar-inertial slam with
  structural constraints,'' in \emph{{Proceedings of the {IEEE} International
  Conference on Robotics and Automation ({ICRA})}}.\hskip 1em plus 0.5em minus
  0.4em\relax IEEE, 2018, pp. 6521--6528.

\bibitem{Hosseinzadeh:etal:ACCV2018}
M.~Hosseinzadeh, Y.~Latif, T.~Pham, N.~Suenderhauf, and I.~Reid, ``Structure
  aware slam using quadrics and planes,'' in \emph{{Proceedings of the Asian
  Conference on Computer Vision ({ACCV})}}, 2018.

\bibitem{Yang:etal:IROS2016}
S.~Yang, Y.~Song, M.~Kaess, and S.~Scherer, ``Pop-up slam: Semantic monocular
  plane slam for low-texture environments,'' in \emph{{Proceedings of the
  {IEEE/RSJ} Conference on Intelligent Robots and Systems ({IROS})}}, 2016.

\bibitem{Yang:Scherer:RAL2019}
S.~Yang and S.~Scherer, ``Monocular object and plane slam in structured
  environments,'' \emph{{{IEEE} Robotics and Automation Letters}}, 2019.

\bibitem{Hosseinzadeh:etal:ICRA2019}
M.~Hosseinzadeh, K.~Li, Y.~Latif, and I.~Reid, ``Real-time monocular
  object-model aware sparse slam,'' in \emph{2019 International Conference on
  Robotics and Automation (ICRA)}.\hskip 1em plus 0.5em minus 0.4em\relax IEEE,
  2019, pp. 7123--7129.

\bibitem{Arndt:etal:IROS2020}
C.~Arndt, R.~Sabzevari, and J.~Civera, ``From points to planes-adding planar
  constraints to monocular slam factor graphs,'' in \emph{{Proceedings of the
  {IEEE/RSJ} Conference on Intelligent Robots and Systems ({IROS})}}, 2020.

\bibitem{Armeni:etal:ICCV2019}
I.~Armeni, Z.~He, J.~Gwak, A.~Zamir, M.~Fischer, J.~Malik, and S.~Savarese,
  ``3d scene graph: A structure for unified semantics, 3d space, and camera,''
  in \emph{{Proceedings of the International Conference on Computer Vision
  ({ICCV})}}, 2019.

\bibitem{Rosinol:etal:ARXIV2020}
A.~Rosinol, A.~Gupta, M.~Abate, J.~Shi, and L.~Carlone, ``{3D Dynamic Scene
  Graphs: Actionable Spatial Perception with Places, Objects, and Humans},''
  \emph{arXiv preprint arXiv:2002.06289}, 2020.

\bibitem{Wald:etal:CVPR2020}
J.~Wald, H.~Dhamo, N.~Navab, and F.~Tombari, ``Learning 3d semantic scene
  graphs from 3d indoor reconstructions,'' in \emph{Proceedings of the IEEE/CVF
  Conference on Computer Vision and Pattern Recognition (CVPR)}, June 2020.

\bibitem{Wu:etal:CVPR2021}
S.-C. Wu, J.~Wald, K.~Tateno, N.~Navab, and F.~Tombari, ``Scenegraphfusion:
  Incremental 3d scene graph prediction from rgb-d sequences,'' in
  \emph{Proceedings of the IEEE/CVF Conference on Computer Vision and Pattern
  Recognition (CVPR)}, June 2021, pp. 7515--7525.

\bibitem{Pearl:book1988}
J.~Pearl, \emph{{Probabilistic reasoning in intelligent systems: networks of
  plausible inference}}.\hskip 1em plus 0.5em minus 0.4em\relax Morgan
  Kaufmann, 1988.

\bibitem{Bishop:Book2006}
C.~M. Bishop, \emph{{Pattern Recognition and Machine Learning}}.\hskip 1em plus
  0.5em minus 0.4em\relax Springer-Verlag New York, Inc., 2006.

\bibitem{Bickson:PhDThesis:2008}
D.~Bickson, ``Gaussian belief propagation: Theory and application,'' Ph.D.
  dissertation, PhD thesis, The Hebrew University of Jerusalem, 2008.

\bibitem{Mur-Artal:etal:TRO2017}
R.~Mur-Artal and J.~D. Tard{\'o}s, ``{ORB-SLAM2: An Open-Source SLAM System for
  Monocular, Stereo, and RGB-D Cameras},'' \emph{{{IEEE} Transactions on
  Robotics ({T-RO})}}, vol.~33, no.~5, pp. 1255--1262, 2017.

\bibitem{Hartley:Zisserman:Book2004}
R.~Hartley and A.~Zisserman, \emph{{Multiple View Geometry in Computer
  Vision}}, 2nd~ed.\hskip 1em plus 0.5em minus 0.4em\relax Cambridge University
  Press, 2004.

\bibitem{Liu:etal:CVPR2019}
C.~Liu, K.~Kim, J.~Gu, Y.~Furukawa, and J.~Kautz, ``{PlaneRCNN: 3D Plane
  Detection and Reconstruction from a Single Image},'' in \emph{{Proceedings of
  the {IEEE} Conference on Computer Vision and Pattern Recognition ({CVPR})}},
  2019.

\bibitem{He:etal:ICCV2017}
K.~He, G.~Gkioxari, P.~Doll{\'a}r, and R.~Girshick, ``Mask r-cnn,'' in
  \emph{{Proceedings of the International Conference on Computer Vision
  ({ICCV})}}, 2017.

\bibitem{Agarwal:etal:ICRA2013}
P.~Agarwal, G.~D. Tipaldi, L.~Spinello, C.~Stachniss, and W.~Burgard, ``Robust
  map optimization using dynamic covariance scaling,'' in \emph{{Proceedings of
  the {IEEE} International Conference on Robotics and Automation ({ICRA})}},
  2012.

\bibitem{Sturm:etal:IROS2012}
J.~Sturm, N.~Engelhard, F.~Endres, W.~Burgard, and D.~Cremers, ``{A Benchmark
  for the Evaluation of {RGB-D} {SLAM} Systems},'' in \emph{{Proceedings of the
  {IEEE/RSJ} Conference on Intelligent Robots and Systems ({IROS})}}, 2012.

\bibitem{CeresManual}
S.~Agarwal, M.~K., and Others, ``Ceres solver,'' \url{http://ceres-solver.org}.

\end{thebibliography}
}

\end{document}